\documentclass[manuscript,screen]{acmart}

\AtBeginDocument{%
  \providecommand\BibTeX{{%
    \normalfont B\kern-0.5em{\scshape i\kern-0.25em b}\kern-0.8em\TeX}}}

\setcopyright{acmlicensed}
\copyrightyear{2018}
\acmYear{2018}
\acmDOI{XXXXXXX.XXXXXXX}

\acmJournal{JACM}
\acmVolume{37}
\acmNumber{4}
\acmArticle{111}
\acmMonth{8}

\acmISBN{978-1-4503-XXXX-X/18/06}

\usepackage[marginal]{footmisc}
\usepackage{tabularx}
\usepackage{multirow}
\usepackage{alphabeta}
\usepackage{amsthm}
\usepackage{amsmath}
\usepackage{booktabs} 
\usepackage{threeparttable}
\usepackage{bbding}
\usepackage{xpatch}
\usepackage{pifont}
\usepackage{wasysym}

\usepackage{amssymb}
\usepackage{svg}
\usepackage{makecell}
\usepackage{amsmath,amsfonts}
\usepackage{algorithmic}
\usepackage{algorithm}
\usepackage{array}
\newcolumntype{C}[1]{>{\centering}p{#1}}
\usepackage[caption=false,font=normalsize,labelfont=sf,textfont=sf]{subfig}
\usepackage{textcomp}
\usepackage{stfloats}
\usepackage{url}
\usepackage{verbatim}
\usepackage{graphicx}
\usepackage{soul}

\newcommand{\cmmnt}[1]{}
\begin{document}

\title{Exploring the Nexus of Large Language Models and Legal Systems: A Short Survey}

\author{Weicong Qin}
\email{qwc@ruc.edu.cn}
\affiliation{%
  \institution{Renmin University of China}
  \streetaddress{Zhongguancun St. 59th}
  \city{Haidian}
  \state{Beijing}
  \country{China}
  \postcode{100872}
}

\author{Zhongxiang Sun}
\email{sunzhongxiang@ruc.edu.cn}
\affiliation{%
  \institution{Renmin University of China}
  \streetaddress{Zhongguancun St. 59th}
  \city{Haidian}
  \state{Beijing}
  \country{China}
  \postcode{100872}
}

\renewcommand{\shortauthors}{Weicong Qin}


\begin{abstract}

With the advancement of Artificial Intelligence (AI) and Large Language Models (LLMs), there is a profound transformation occurring in the realm of natural language processing tasks within the legal domain. The capabilities of LLMs are increasingly demonstrating unique roles in the legal sector, bringing both distinctive benefits and various challenges. This survey delves into the synergy between LLMs and the legal system, such as their applications in tasks like legal text comprehension, case retrieval, and analysis. Furthermore, this survey highlights key challenges faced by LLMs in the legal domain, including bias, interpretability, and ethical considerations, as well as how researchers are addressing these issues. The survey showcases the latest advancements in fine-tuned legal LLMs tailored for various legal systems, along with legal datasets available for fine-tuning LLMs in various languages. Additionally, it proposes directions for future research and development.
\end{abstract}


\maketitle





\section{Introduction}
\label{sec:intro}
The emergence of large language models (LLMs) has revolutionized numerous domains, spanning from natural language processing (NLP)\cite{brown2020language,touvron2023llama} to computer vision\cite{wu2023visual,shao2023prompting}, and even extending to reinforcement learning~\cite{du2023guiding}. 

Notably, LLMs are increasingly being applied in legal text processing and understanding, where they perform a variety of tasks. These tasks include predicting legal judgments, reasoning with statutes, analyzing privacy policies, and generating summaries of legal cases~\cite{trautmann2022legal, yu2022legal, chalkidis2023chatgpt, deroy2023ready, blair2023can, savelka2023explaining, nguyen2023black, tang2023policygpt, nguyen2024employing}. LLMs are also being used to improve legal case retrieval and analysis, providing advice on specific cases and drafting legal documents~\cite{macey2023chatgpt, iu2023chatgpt, zhang2023reformulating, wu2023precedent, zhou2023boosting, sun2024logic}. While these models show promise, they require domain-specific training and human oversight to ensure accuracy and reliability in legal applications. The integration of LLMs in the legal field is seen as a complementary tool to legal professionals, enhancing efficiency and innovation while maintaining the need for human expertise.

However, the assimilation of LLMs into legal practices has precipitated a host of legal problems, encompassing privacy encroachments, biases, and interpretability issues~\cite{tamkin2021understanding, huang2023survey, grossman2023gptjudge, deroy2023questioning, felkner2022towards, abid2021persistent, li2023dark, lindholm2022machine}. This survey embarks on an exploration of the incorporation of LLMs within the legal domain. It delves into the manifold applications of LLMs in legal tasks, scrutinizes the legal hurdles that ensue from their deployment, and investigates the reservoirs of data that can be harnessed to tailor LLMs for legal contexts~\cite{xiao2018cail2018,zheng2021does,ma2021lecard}. Lastly, it contemplates several promising directions and concludes on these observations. Through this endeavor, we aspire to furnish an overview of the contemporary status of LLMs in the legal domain, shedding light on both their potential advantages and impediments.

\subsection{Pertinent Surveys}

The extant corpus of survey literature concerning intelligent legal systems predominantly fixates on conventional natural language technologies. Some of these surveys concentrate solely on a singular legal task, such as prognosticating legal case outcomes. Conversely, others encompass a spectrum of legal tasks. The majority of these survey articles summarize current resources, including open-source utilities and datasets, pertinent to legal research.

\begin{table}[!htp]
    \centering
    \caption{Comparison with existing surveys. For each survey, we summarize the topics covered and the main scope to survey.}
    \scalebox{1.0}{
\begin{tabular}{ll|l|l|l|l|l|l}
\hline
\multicolumn{2}{l|}{Surveys}                                                                   & \cite{chalkidis2019deep} & \cite{cui2022survey} & \cite{dias2022state} & \cite{katz2023natural} & \cite{sun2023short} & ours       \\ \hline
\multicolumn{1}{l|}{\multirow{7}{*}{Topics}} & LLMs' applications                              & $\times$                 & $\times$             & $\times$             & $\times$               & \checkmark          & \checkmark \\
\multicolumn{1}{l|}{}                        & \quad --Categorized for legal tasks                 & $\times$                 & $\times$             & $\times$             & $\times$               & $\times$            & \checkmark \\ \cline{2-8} 
\multicolumn{1}{l|}{}                        & Fine-tuned LLMs for various countries and regions & $\times$                 & $\times$             & $\times$             & $\times$               & $\times$            & \checkmark \\ \cline{2-8} 
\multicolumn{1}{l|}{}                        & Multi-domain                                    & \checkmark               & $\times$             & \checkmark           & \checkmark             & \checkmark          & \checkmark \\ \cline{2-8} 
\multicolumn{1}{l|}{}                        & Dataset sources                                 & \checkmark               & \checkmark           & \checkmark           & \checkmark             & \checkmark          & \checkmark \\
\multicolumn{1}{l|}{}                        & \quad--Large-scale sources specifically for LLMs   & $\times$                 & $\times$             & $\times$             & $\times$               & \checkmark          & \checkmark \\
\multicolumn{1}{l|}{}                        & \quad\quad --Multi-language                            & $\times$                 & $\times$             & $\times$             & $\times$               & $\times$            & \checkmark \\ \hline
\multicolumn{2}{l|}{Latest year}                                                               & 2019                     & 2022                 & 2022                 & 2023                   & 2023                & 2024       \\ \hline
\end{tabular}
}
    \label{tab:survey_analysis}
\end{table}


As shown in Table~\autoref{tab:survey_analysis}, there is a lack of surveys specifically targeting large language models in the legal domain, apart from~\cite{sun2023short}. Even though ~\cite{sun2023short} fills this gap, its content is limited by the development of LLMs in the legal domain as of March 2023 and needs further enrichment. Therefore, this work, building upon prior research, aims to conduct a comprehensive investigation and analysis of the latest LLMs-based intelligent legal systems, covering multiple countries and regions, as well as multiple languages, to provide an exhaustive and methodical survey.
The classification of the surveyed papers are shown in~\autoref{tab:paper_classification}.

\begin{table}[!htp]
  \centering
  \caption{Classification of papers}
  \label{tab:paper_classification}
      \scalebox{1.0}{
  \begin{tabular}{ll}
    \toprule
    \textbf{Papers} & \textbf{Category}\\
    \midrule
    \cite{trautmann2022legal,yu2022legal,chalkidis2023chatgpt,deroy2023ready,blair2023can,savelka2023explaining,nguyen2023black,tang2023policygpt,nguyen2024employing,macey2023chatgpt,iu2023chatgpt,zhang2023reformulating}
 & \multicolumn{1}{l}{\multirow{2}{*}{Applications of Large Language Models in Legal Tasks}}\\
\cite{wu2023precedent,zhou2023boosting,sun2024logic,nay2022law,dal2023legal,sileno2023text,nay2023large,simmons2023garbage,westermann2023llmediator}& \multicolumn{1}{l}{}  \\ \hline
\cite{zhang2023reformulating, Lawyer-LLama, cui2023chatlaw, yue2023disclawllm, lee2023lexgpt, qasem2023towards, gesnouin2024llamandement}  & Fine-tuned Large language models in Various Nations and
Regions\\ \hline
    \cite{tamkin2021understanding, huang2023survey, grossman2023gptjudge, deroy2023questioning, li2023dark, lindholm2022machine, StochasticParrots, felkner2022towards, abid2021persistent, cheong2023us}
 & Legal Problems of Large Language Models\\ \hline
    \cite{xiao2018cail2018, ma2021lecard, li2023lecardv2, chalkidis2023lexfiles, bauer2023legal, zheng2021does, ostling2024cambridge, niklaus2023multilegalpile} & Data Resources for Large Language Models in Law\\ 
    \bottomrule
  \end{tabular}
  }
\end{table}

\subsection{Contributions}



In this survey, we have presented several advancements in the intersection of law and natural language processing, including:
\begin{itemize}
 \item Providing an overview of how LLMs are applied in legal tasks such legal Text Processing and Understanding, legal Case Retrieval and Analysis, and so on.
 \item Introduced fine-tuning legal LLMs tailored for the legal systems of various countries and regions.
 \item Conducting an analysis of the legal issues arising from the utilization of large language models in law, covering concerns related to privacy, bias, fairness, and the unique challenges of LLMs, such as the phenomenon of hallucination and stochastic parrots.
 \item Discussing the available large-scale data resources in various languages for refining LLMs in the legal domain.
\item Proposing avenues for future research to tackle the legal hurdles associated with employing large language models in law, such as developing methodologies to mitigate bias and ensure transparency.
\end{itemize}

Through these contributions, we aim to offer a comprehensive understanding of the current landscape of large language models in law, emphasizing both their potential advantages and challenges. Our objective is to stimulate further investigation in this domain and support the responsible and ethical integration of large language models into legal practice. For an updated list of references, please refer to our paper repository at \url{https://github.com/Jeryi-Sun/LLM-and-Law}.

\section{Applications of Large Language Models in Legal Tasks}

This chapter explores the latest advancements of Large Language Models (LLMs) like ChatGPT in the realm of legal tasks. It underscores their transformative potential within the legal field, examines the challenges and opportunities they encounter, and elucidates the intricate interplay and differences among related research efforts.

\subsection{Legal Text Processing and Understanding}

In the dynamic field of Legal Text Processing and Understanding, an array of compelling studies have come to the forefront, concentrating on the utilization of LLMs to navigate the complex landscape of legal documentation. These papers explore the intricate processes of legal judgment prediction, statutory reasoning, legal text entailment, privacy policy analysis, and the generation of legal case summaries. As we delve into these studies, we gain a deeper appreciation for the nuanced capabilities of LLMs in interpreting and generating legal text, tasks that are pivotal to the advancement of legal technology and the enhancement of legal practice.

Regarding specific legal downstream tasks, the study by~\cite{trautmann2022legal} serves as a starting point, where the concept of Legal Prompt Engineering (LPE) is introduced. LPE demonstrates the potential of zero-shot performance in applying general-purpose LLMs to the legal domain without domain-specific data, revealing the ability of LLMs to transfer their learned skills to specialized sectors, albeit with room for improvement when compared to supervised approaches.
Building on the premise of prompt engineering, \cite{yu2022legal} further refines this method by employing Chain-of-Thought (CoT) prompts and legal reasoning techniques such as IRAC. This paper underscores the effectiveness of domain-specific prompting in enhancing legal reasoning capabilities, achieving significant accuracy improvements in legal entailment tasks, and marking a step forward in the evolution of LLM application in the legal domain.
The progress in prompt-based methods leads us to explore the adaptability of conversational LLMs in the legal field. The paper by~\cite{chalkidis2023chatgpt} reflects on the performance of OpenAI's ChatGPT model. While ChatGPT exhibits an impressive understanding of legal documents, outperforming baseline models, it still falls short in comprehensive legal benchmarks, suggesting that conversational models, while promising, require further specialization for legal applications.
The specialization of LLMs is further scrutinized in~\cite{deroy2023ready} which evaluates the readiness of both domain-specific and general-domain models for summarizing legal judgments. Despite the advantages of abstractive summarization models in generating coherent summaries, their tendency to introduce errors necessitates a human-in-the-loop approach, highlighting the current limitations of these technologies in legal document processing. 
Furthermore,~\cite{blair2023can} evaluates GPT-3's capability in statutory reasoning, using the SARA dataset to test its understanding and application of legal statutes. The study finds that GPT-3, while surpassing previous benchmarks, struggles with imperfect knowledge of actual laws and reasoning about novel legal content, highlighting the need for further advancements in AI's legal reasoning abilities.
To enhance the factual accuracy of LLM outputs,~\cite{savelka2023explaining} presents an augmented approach where LLMs are supplemented with relevant case law context. This augmentation significantly reduces hallucinations and increases the factual accuracy of legal term explanations, showcasing the value of context-aware systems in the legal domain.
The evolutionary trajectory of LLMs is then examined in~\cite{nguyen2023black} which analyzes the performance of different generations of GPT models on a Japanese legal textual entailment task. The study emphasizes the importance of black-box analysis in understanding how these models process legal texts and identifies the impact of training data distribution on performance, setting the stage for future enhancements.
\cite{tang2023policygpt} turns the focus to the practical application of LLMs in analyzing verbose privacy policies. PolicyGPT's impressive performance in classifying text segments demonstrates the efficacy of LLMs in streamlining complex legal text analysis, surpassing traditional machine learning models and representing a significant stride towards automating legal text comprehension.
Lastly,~\cite{nguyen2024employing} introduces an innovative approach that combines ChatGPT's capabilities with label models to improve legal text entailment performance. By treating ChatGPT's provisional answers as noisy predictions, this method refines the model's outputs, achieving a marked improvement in accuracy and setting a new state-of-the-art benchmark.

In summary, these papers collectively underline the rapid progress and potential of large language models in processing and understanding legal texts, from multilingual judgment prediction to privacy policy analysis. However, they also highlight the challenges ahead, particularly in terms of enhancing the accuracy, reliability, and domain specificity of these models. As this body of work demonstrates, while large language models are approaching levels of performance that can assist and even automate certain legal tasks, a nuanced approach that includes domain-specific prompting, contextual augmentation, and human oversight is still essential for their effective implementation in the legal field. The path forward will likely involve a symbiotic relationship between AI and legal experts, leveraging the strengths of each to advance the field of legal text processing and understanding.

\subsection{Legal Case Retrieval and Analysis}

In the realm of Legal Case Retrieval and Analysis, an emerging frontier in the intersection of law and technology, a collection of insightful studies shed light on the innovative application of LLMs. These scholarly works delve into the intricacies of harnessing LLMs for not only retrieving and analyzing legal cases but also for articulating the rationale behind case retrievals through the application of logic rules. Through a series of papers that examine the capabilities, frameworks, and applications of LLMs in legal settings, we gain insights into how these tools can revolutionize the field.

Beginning with the implementation of AI in practical legal advice, the study of~\cite{macey2023chatgpt} as quasi-expert legal advisors showcases the application of ChatGPT in providing advice for a specific case – that of Tom Hayes and the Libor scandal. This paper demonstrates the potential for LLMs to generate quality legal advice, suggesting that such systems could soon play a central role in the legal profession, challenging the traditional role of lawyers. The study exemplifies how AI, by leveraging vast databases of legal case law and analysis, can offer nuanced perspectives on complex legal issues, potentially leading to calls for retrials based on new insights into case evidence.
In the realm of litigation, \cite{iu2023chatgpt} evaluates ChatGPT's capabilities in drafting legal documents and providing legal advice. While the system shows promise in understanding and articulating legal claims, it currently falls short in areas such as identifying recent case law. Thus, ChatGPT is positioned as a supplementary tool to litigation lawyers rather than a replacement, highlighting the importance of human expertise in navigating the dynamic legal landscape.
Addressing the issue of domain specificity,~\cite{zhang2023reformulating} introduces an adapt-retrieve-revise framework that effectively prevents the generation of inaccurate content by continuing the training of a smaller LLM on domain-specific data. This technique not only improves the accuracy of the model but also underscores the need for LLMs to be fine-tuned to specialized domains, such as Chinese law, to avoid generating irrelevant or erroneous content.
Furthering the concept of domain-specific adaptation,~\cite{wu2023precedent} presents a system that marries the comprehensive language understanding of LLMs with the precision of domain models. This hybrid approach, which leverages precedents for Legal Judgment Prediction (LJP), demonstrates how LLMs can be enriched with domain-specific insights to deliver more accurate predictions in legal contexts.
Moving to the enhancement of legal case retrieval, the study by~\cite{zhou2023boosting} focuses on refining the retrieval process by employing LLMs to distill salient content from lengthy, complex legal queries. This technique of content selection and summarization by LLMs significantly improves the performance of both sparse and dense retrieval models, highlighting the importance of understanding and extracting relevant legal-specific elements for effective case retrieval.
Finally, the study by~\cite{sun2024logic} confronts the need for interpretable and faithful explanations within legal case retrieval systems. The proposed Neural-Symbolic enhanced Legal Case Retrieval (NS-LCR) framework integrates case-level and law-level logic rules into the retrieval process, furnishing users with cogent explanations for the retrieval results. This neuro-symbolic approach not only boosts retrieval effectiveness but also provides the much-needed transparency for legal decision-making.

In summary, the surveyed papers collectively underscore the transformative impact LLMs like GPT-4 can have on the field of legal case retrieval and analysis. From augmenting the quality of legal advice to enhancing the accuracy of legal judgment predictions, LLMs hold the promise of streamlining legal processes while ensuring the provision of clear, logical explanations for legal professionals. Yet, these advancements come with caveats, such as the need for domain-specific training and the recognition of AI as a complementary tool rather than a wholesale replacement for human expertise. The continued development and integration of LLMs in legal settings present a two-fold benefit: they introduce efficiency and innovation while urging the legal profession to adapt to a new era of technology-driven legal service.

\subsection{Legal Education and Examinations}


In the realm of legal education and examinations, the integration of large language models has sparked a revolutionary dialogue about the potential transformation of traditional methodologies. It is in this context that we examine the role of LLMs in engaging with the academic and pedagogical aspects of law. The papers under review explore this theme from the perspectives of both student evaluation and faculty assistance.

An intriguing experiment documented in~\cite{choi2023chatgpt} where ChatGPT was subjected to law school final exams consisting of multiple-choice and essay questions. This study not only showcases the capacity of AI to mimic the average law student, managing a passing C+ grade across four courses, but also ignites a conversation on how legal education might adapt to incorporate AI as a tool for learning and evaluation. The findings suggest that while ChatGPT can produce work that meets basic legal educational standards, there is an inherent limitation in the depth and nuance of understanding that is required for higher-level legal analysis. This study serves as a foundational reference point, demonstrating that AI's ability to engage with complex legal problems is nascent but promising.

Additionally,~\cite{hargreaves2023words} evaluates the risk of ChatGPT to academic integrity in higher education by having the AI model generate answers to a variety of law school exams from a common law jurisdiction.~\cite{hargreaves2023words} found that while ChatGPT excelled at essay-based questions discussing international legal instruments and general legal principles, it struggled with problem-style questions that required applying specific factual scenarios to local legislation or jurisprudence. This suggests that traditional law school assessments are currently somewhat resistant to the challenges posed by ChatGPT, although this may change as AI technology evolves. Rather than blocking AI use, the research proposes a proactive approach to educating students on the ethical and appropriate integration of AI into their learning and assessment processes, acknowledging the transformative potential of AI technologies like ChatGPT on university teaching and evaluation methodologies.

Building on the initial insights provided,~\cite{pettinato2023chatgpt} pushes the boundaries of AI's capabilities further into the realm of academic facilitation. By testing ChatGPT's proficiency in executing tasks commonly handled by law professors, including service-related and teaching-related prompts, the study reveals that AI can not only assist in drafting near-complete work for routine tasks but also provide a substantive base for more intricate assignments. This duality of function emphasizes the versatility of large language models and supports the thesis that AI can be an asset in legal education by alleviating some of the workload for professors, thereby allowing them to focus on the more critical and nuanced aspects of legal teaching.

The synergy between these two pieces of research paints a comprehensive picture of how AI, particularly ChatGPT, can be integrated into legal education and examination systems. We see a landscape where AI can provide basic educational support and alleviate administrative burdens, creating space for deeper intellectual engagement in legal studies.
As we consider the incorporation of large language models into legal education and examination practices, it is clear that AI can play a supportive role, though it is not without its limitations.~\cite{choi2023chatgpt} and~\cite{pettinato2023chatgpt} collectively highlight the opportunities for AI to contribute to the academic sphere while simultaneously challenging educators to reassess and evolve the current paradigms of teaching and evaluating law students. This duality suggests a future where AI's role is not to replace but to enhance the legal educational experience, offering new avenues for student support and academic assistance while upholding the rigorous standards of legal scholarship.

\subsection{Legal Practice and Assistive Tools}


In the realm of legal practice, the integration of LLMs as assistive tools signifies a transformative shift, wherein the traditional processes are being redefined and enhanced through AI interventions. Many researchers delve into distinct yet interconnected aspects of this paradigm, from understanding legal standards and summarizing judicial decisions to structuring legal texts and mediating disputes. These studies collectively paint a picture of a future where LLMs are integral to legal processes, offering both efficiency and new forms of interaction between legal professionals and the law itself.

In the quest to harmonize artificial intelligence with the intricate fabric of human values and societal norms, the paper of~\cite{nay2022law} proposes a legal informatics approach to align AI with human values through the "Law Informs Code" agenda, embedding legal reasoning into AI. It highlights the challenge of precisely directing AI behavior due to the inherent unpredictability in legal and societal contexts. The research suggests using data from legal processes to better define human goals for AI, enhancing alignment and utility. 
~\cite{dal2023legal} examines how tax judges and lawyers can leverage digital technology in their practice. The use of LLMs, particularly GPT-4, in creating summaries of judicial decisions is not only about facilitating comprehension but also extracting pertinent information such as legal issues and decision-making criteria. The practical application of these summarizations and the development of a prototype application suggest a movement towards accessible and user-friendly digital legal tools. The satisfactory evaluation by legal experts confirms the relevance of LLMs in handling complex legal texts, setting a foundational understanding of the role of AI in legal documentation.
Building upon the premise of document handling,~\cite{sileno2023text} explores the transformation of legislative text into structured representations. The utilization of LLMs to encode legislative information points to a future where the bottleneck of manual analysis is alleviated, accelerating the creation of legal expert systems. With a significant portion of generated pathways being as good as or better than manual ones, the study highlights the potential of LLMs in streamlining the development of tools that help laypeople understand legal applications, enhancing accessibility to legal knowledge.
Moving to the aspect of legal reasoning and the operational performance of LLMs, the research of ~\cite{nay2023large} underscores the challenges in goal specification for AI and the potency of legal standards as a means of robust communication. The empirical study assessing LLMs’ understanding of fiduciary obligations reflects a significant improvement in AI's grasp of legal concepts, with OpenAI's latest LLM showing substantial accuracy. This progression suggests a future where LLMs can interpret and act within the spirit of legal directives, promising more nuanced AI participation in legal decision-making processes.
In contrast, \cite{simmons2023garbage} presents a practical application of LLMs in crime detection using zero-shot reasoning from textual descriptions of videos. The paper draws attention to the current limitations of automated video-to-text conversion, highlighting the need for quality inputs to harness the full capabilities of LLMs in legal contexts. This study emphasizes the importance of data quality in legal AI applications and points towards future improvements in multimodal AI capabilities.
Lastly,~\cite{westermann2023llmediator} showcases the application of LLMs in the human-centric aspect of legal practice: dispute resolution. The introduction of LLMediator hints at a future where AI not only assists but potentially autonomously engages in negotiations and mediation, demonstrating an initial step towards AI-facilitated conflict resolution. This aligns with the broader theme of AI as an assistive tool, contributing to more efficient and cooperative dispute resolution methods.

In conclusion, these papers collectively underscore the expanding capabilities of LLMs in legal practice, presenting a multifaceted view of AI as a transformative force in law. From summarizing and structuring documents to understanding legal standards and aiding in dispute resolution, LLMs are paving the way for more nuanced, efficient, and accessible legal processes. The continuous improvement in the understanding and application of LLMs in legal standards suggests an evolution towards more sophisticated AI tools that can integrate seamlessly into various facets of legal practice. Nevertheless, the research also signals the necessity for high-quality data and the cultivation of shared understandings of legal concepts between humans and AI. Future studies and developments must aim to bridge these gaps, ensuring that the deployment of LLMs in legal practice is as robust, reliable, and ethically aligned as the profession demands.

\subsection{Chapter Summary}

Collectively, these papers contribute to our understanding of domain adaptation and model improvement in AI for legal applications. They span from theoretical frameworks that argue for embedding legal reasoning within AI systems to practical methodologies that seek to adapt AI models to specific legal domains.

\section{Fine-tuned Large language models in Various Countries and Regions}



Although general-purpose LLMs exhibit considerable performance in legal domains, they fail to meet the demands of more specialized legal tasks. As a result, researchers have turned their attention to fine-tuning large models specifically for legal applications. The fine-tuning of large language models for legal-specific domains has shown immense potential in addressing tasks related to legal texts. Research in this field aims to enhance the models' understanding of legal language, thereby providing more precise and effective tools and methods for legal practice, research, and decision support.

\subsection{Mainland China}

To advance the utilization of large language models (LLMs) within the legal domain of Mainland China, a progressive approach has been undertaken, evolving from basic fine-tuning methodologies to more sophisticated integrations of external knowledge sources and specialized retrieval mechanisms.

Initially, LawGPT\footnote{https://github.com/pengxiao-song/LaWGPT} laid the groundwork by enhancing general Chinese LLMs with legal domain-specific vocabularies and extensive pre-training on Chinese legal corpora. This foundation facilitated the construction of dialogue question-and-answer datasets sourced from legal contexts, refining the model's comprehension and application of legal content through fine-tuning processes.

Subsequently, Lawyer LLaMA\footnote{https://github.com/AndrewZhe/lawyer-llama} introduced a framework emphasizing continual training and the incorporation of domain knowledge to impart professional skills to LLMs~\cite{Lawyer-LLama,lawyer-llama-report}. Notably, a retrieval module was integrated to counteract hallucination issues, ensuring generated responses remained grounded in relevant legal articles.

Building upon these advancements, LexiLaw\footnote{https://github.com/CSHaitao/LexiLaw} merged general and legal domain datasets for fine-tuning, recognizing the necessity of data diversity to prevent overfitting while enriching the model's expertise. This amalgamation encompassed general domain data, legal documents, laws and regulations, legal reference books, and legal Q\&A datasets, augmenting the model's understanding and proficiency in legal contexts.

Further innovation emerged with ChatLaw\footnote{https://github.com/PKU-YuanGroup/ChatLaw}~\cite{ChatLaw,cui2023chatlaw}, which combined external knowledge bases with vector database retrieval and keyword screening techniques to enhance the accuracy of legal LLMs. The integration of consult, reference, self-suggestion, and response modules facilitated comprehensive legal consultations while mitigating the risk of erroneous outputs.

DISC-LawLLM\footnote{https://github.com/FudanDISC/DISC-LawLLM} introduced intelligent legal systems leveraging LLMs with refined reasoning capabilities and an integrated retrieval module for accessing external legal knowledge~\cite{yue2023disclawllm}, using a dataset constructed with legal syllogism prompting strategies to simulate various legal scenarios and tasks. Benchmarking against the DISC-Law-Eval dataset, this system showcased superior performance in legal knowledge and reasoning tasks, underscoring its efficacy in diverse legal scenarios and user contexts.

 In terms of efficiency, ~\cite{zhang2023reformulating} recognizes the shortcomings (inefficiency and inaccuracy) of general large language models, such as GPT-4, when it comes to domain-specific tasks like Chinese law. To address this, they propose a novel framework which involves adapting a smaller, 7B language model through continued learning on domain-specific data. This adapted model then generates draft answers which are improved by retrieving evidence from an external knowledge base and using GPT-4's capabilities to revise and finalize the answers. The process illustrates an innovative adaptation mechanism that harnesses both the efficiency of smaller models and the sophisticated revision capabilities of larger ones, significantly improving accuracy in specific legal tasks.

 In short, the evolution of fine-tuned large models within Mainland China's legal domain reflects a trajectory from basic fine-tuning approaches to sophisticated integrations of external knowledge sources, retrieval mechanisms, and reasoning capabilities. These advancements signify a paradigm shift towards more comprehensive and effective utilization of LLMs in legal contexts, promising enhanced accuracy, reliability, and accessibility for legal practitioners and stakeholders.

\subsection{Taiwan}
Regarding the laws of Taiwan, \cite{lee2023lexgpt} presents LexGPT 0.1, a specialized language model for the legal domain based on GPT-J models, pre-trained with the Pile of Law dataset, aiming to assist legal professionals without the need for programming skills. It explores the "No Code" approach for fine-tuning models for downstream tasks like classification and discusses the potential for future enhancements in legal domain applications.

\subsection{Palestine}
Regarding Palestinian laws, \cite{qasem2023towards} presents a cooperative-legal question-answering LLM-based chatbot focused on Palestinian laws. It compares the chatbot's auto-generated answers to those crafted by a legal expert, evaluating its performance with 50 queries. Results show an overall accuracy rate of 82\% and an F1 score equivalent to 79\% in answering the queries.

\subsection{France}
For French law, ~\cite{gesnouin2024llamandement} presents LLaMandement, a cutting-edge large language model developed in collaboration with the French government to enhance the processing of parliamentary sessions. It specializes in generating impartial summaries of legislative proposals, thereby improving the efficiency of document production for interministerial meetings.

\subsection{Chapter Summary}

In conclusion, the evolution of fine-tuned large models within legal domains across Mainland China, Taiwan, Palestine, and France signifies a global trend towards tailored solutions for specialized legal tasks. These initiatives emphasize integrating external knowledge sources and domain-specific training, reflecting a shift towards more nuanced and effective utilization of large language models in legal contexts.



Throughout this chapter, the versatile applications and research avenues of LLMs in the legal domain are evident, spanning from text processing and case analysis to educational aids and real-world legal applications. Furthermore, they extend to theoretical inquiries into law and AI integration, as well as model enhancements. These collective efforts propel the advancement of legal AI, enriching the efficacy and caliber of legal services provided.


\section{Legal Problems of Large Language Models} 

When discussing the limitations and issues of large models, we inevitably face a series of complex and significant issues. Despite the impressive technical capabilities of Large Language Models (LLMs) in generating text, they may not possess a complete understanding of the underlying meaning of language~\cite{tamkin2021understanding}. This leads to a range of problems, including but not limited to mimicry without comprehension, false correlations, a lack of understanding of causality, and the potential presence of biases. Additionally, LLMs may fall into hallucinations, generating content that is unrelated or illogical based on the input data~\cite{huang2023survey}. In their application within the judicial system, these issues could pose serious challenges and risks~\cite{grossman2023gptjudge,deroy2023questioning}, including misinterpretation and distortion of evidence, as well as impacts on fundamental values. Therefore, it is essential to delve into these issues and seek effective strategies to ensure that the use of large language models does not adversely affect our values and legal systems.

\subsection{Stochastic Parrots and Spurious Correlations}

Stochastic Parrots, as metaphorically described in the context of artificial intelligence, depict systems that generate seemingly intelligent outputs without genuine comprehension or conscious intent, akin to parrots mimicking human speech without understanding its meaning~\cite{StochasticParrots}. For example, in the process of generating or analyzing legal documents, if a LLM relies solely on surface patterns in the training data rather than understanding legal concepts or language meanings, it may produce misleading results or inferences. This could lead to erroneous legal advice, inaccurate legal documents, or unfair decisions, thereby negatively impacting the judicial system and citizens' rights.

~\cite{li2023dark} delves into the phenomenon of Stochastic Parrots within LLMs, shedding light on their tendency to imitate language without true comprehension, as well as their limitations in grasping causal relationships~\cite{tamkin2021understanding,lindholm2022machine}. Additionally,~\cite{StochasticParrots} suggests LLMs often rely on probabilistic processes learned from data, generating outputs that appear coherent but lack genuine understanding of language or context. Studies have shown that these models essentially repeat back contents of data with added stochasticity, indicating a mimicry of language rather than comprehension~\cite{lindholm2022machine}. Consequently, despite their ability to produce seemingly intelligent responses, LLMs may struggle to discern incorrect, out-of-context, or socially inappropriate information~\cite{lindholm2022machine}.

Furthermore, LLMs' understanding of causal relationships remains questionable, as they may rely on spurious correlations within data rather than comprehending underlying causality~\cite{tamkin2021understanding}. This limitation hampers their ability to discern meaningful patterns and make accurate predictions beyond mere data replication~\cite{tamkin2021understanding}. Consequently, the potential ramifications of LLMs' inability to grasp causal relationships pose ethical and legal challenges, necessitating a deeper understanding and regulatory scrutiny~\cite{li2023dark}.

\subsection{Biases and Hallucination in Legal Aspect}

\subsubsection{(1) Biases}

The prevalence of biases within LLMs poses significant concerns within legal contexts~\cite{deroy2023questioning}. These biases, spanning various dimensions such as racial, gender, religious, and LGBTQ+ biases, as highlighted in recent studies~\cite{felkner2022towards,abid2021persistent,tamkin2021understanding}, can profoundly impact legal decision-making processes. 

These biases manifest in the outputs generated by LLMs, affecting downstream tasks and potentially reinforcing harmful stereotypes and prejudices. For instance, research has shown persistent anti-Muslim bias in LLMs like GPT-3, with Muslim frequently associated with violence and terrorism~\cite{abid2021persistent}. Similarly, biases against queer and transgender individuals have been observed in models like BERT, leading to homophobic and transphobic outputs~\cite{felkner2022towards}. In addition,~\cite{deroy2023questioning} finds that Gender-related keywords, Race-related keywords, Keywords related to crime against women, Country names and religious keywords may induce biases in the generation of case judgment summaries. Understanding and addressing these biases are essential for ensuring the ethical and equitable deployment of LLMs in various applications.

Several studies have shed light on the presence of biases in LLMs and proposed approaches to mitigate them. For instance, one approach involves fine-tuning models on data representing underrepresented or marginalized groups, as suggested in the study addressing anti-queer bias~\cite{felkner2022towards}. By finetuning models on a natural language corpus written by LGBTQ+ individuals, researchers were able to mitigate homophobic bias in BERT. Additionally, adversarial text prompts have been proposed as a means to reduce bias in LLM outputs~\cite{abid2021persistent}. By introducing positive distractions through carefully crafted prompts, researchers were able to decrease the occurrence of violent associations with Muslims in model outputs. However, these approaches are just the beginning, and further cross-disciplinary collaboration is necessary to develop comprehensive strategies for managing biases in LLMs~\cite{tamkin2021understanding}. Moreover, there is a pressing need for ongoing research to understand the root causes of biases in LLMs and their implications for society.

\subsubsection{(2) Hallucination}

Hallucination occurs when LLMs generate content that deviates from user input, context, or established knowledge~\cite{huang2023survey}. This phenomenon introduces significant uncertainties into legal proceedings, where accuracy and reliability are paramount. 

Several studies shed light on the legal implications of LLM hallucination. For instance, research~\cite{li2023dark} warns that current AI regulatory paradigms, such as those in the EU, might underestimate the risks posed by new LLMs. Furthermore,~\cite{huang2023survey} categorizes hallucinations into various types, including input-conflicting and fact-conflicting, emphasizing the complexity of addressing such distortions within legal contexts. These findings underscore the urgent need for legal systems to adapt and develop strategies to mitigate the impact of hallucination in LLM-generated content.

\subsection{The Challenge to Legal Systems and Fundamental Values}

In an era where LLMs blur the lines between human and machine-generated content, the legal landscape faces unprecedented challenges. The proliferation of AI-generated evidence, as discussed in~\cite{grossman2023gptjudge}, poses a significant hurdle in distinguishing authentic from fake evidence, thereby complicating litigation procedures and potentially overwhelming the courts with AI-generated lawsuits. Moreover, biases inherent in AI-generated case summaries, as explored in~\cite{deroy2023questioning}, raise concerns about fairness and accuracy in legal decision-making processes. These challenges not only undermine the integrity of the judicial system but also threaten fundamental human values such as autonomy, privacy, and equality, as highlighted in~\cite{cheong2023us}.

To address these challenges, it is imperative to align model objectives with human values, prioritizing fairness and accountability in AI systems, as suggested in~\cite{tamkin2021understanding}. Furthermore, proactive measures such as evolving legal frameworks and interdisciplinary collaborations are essential to mitigate the risks posed by generative AI to fundamental human values, as advocated in~\cite{cheong2023us}. By adopting practical recommendations outlined in~\cite{grossman2023gptjudge}, courts and attorneys can navigate the complexities of AI-generated evidence and ensure the integrity of legal proceedings in an increasingly AI-driven world. Additionally, ongoing research and scrutiny into biases within AI-generated content, as emphasized in~\cite{deroy2023questioning}, are crucial to safeguarding against discriminatory outcomes and upholding principles of justice and equality.

\subsection{Chapter Summary}

In conclusion, as generative AI continues to reshape the legal landscape, it is imperative to address the challenges it poses to the judicial system and fundamental human values. By acknowledging these challenges and implementing proactive measures, the legal community can adapt to the evolving technological landscape while upholding the principles of justice, fairness, and equality.

\section{Data Resources for Large Language Models in Law}

As researchers become increasingly interested in using large models for legal tasks, their demand for specialized legal data is also growing. This is because the specialized nature of legal language and knowledge requires specific data to effectively adapt and fine-tune LLMs.

\subsection{Chinese}

One essential data resource is the CAIL2018 dataset, introduced in CAIL2018~\cite{xiao2018cail2018}. Comprising more than 2.6 million criminal cases from China, this dataset allows researchers to delve into various aspects of LJP, such as multi-label classification, multi-task learning, and explainable reasoning. The detailed annotations of applicable law articles, charges, and prison terms provide a rich source of information for LLMs to specialize in the legal domain.

Furthermore, LeCaRD offers a novel dataset for legal case retrieval based on the Chinese law system~\cite{ma2021lecard}. The Chinese Legal Case Retrieval Dataset (LeCaRD) consists of 107 query cases and over 43,000 candidate cases sourced from criminal cases published by the Supreme People's Court of China. This dataset, along with the relevance judgment criteria and query sampling strategy proposed in the paper, provides a valuable resource for specializing LLMs in the Chinese legal system and its unique terminology, logic, and structure.
Building upon its predecessor, version 1, LeCaRDv2 is introduced~\cite{li2023lecardv2}. With 800 queries and 55,192 candidates from 4.3 million criminal case documents, it offers extensive coverage of criminal charges, enriching relevance criteria with characterization, penalty, and procedure aspects. Employing a two-level candidate set pooling strategy, it effectively identifies potential candidates for each query, all annotated by multiple criminal law experts, ensuring accuracy. 

An increasing number of Chinese legal researchers are dedicating themselves to curating and providing more high-quality open-source legal data\footnote{https://github.com/pengxiao-song/awesome-chinese-legal-resources}. 
This commitment to enhancing open-source legal data contributes substantially to the development and refinement of legal big language models, empowering them with broader, more diverse datasets to better understand and navigate complex legal language and contexts.

\subsection{English}

\cite{chalkidis2023lexfiles} proposes LeXFiles, which is a comprehensive English legal dataset encompassing over 19 billion tokens from six major English-speaking legal systems, including EU, CoE, Canada, US, UK, and India. It consists of 11 distinct sub-corpora covering legislation, case law, and contracts to facilitate the development and analysis of legal-oriented language models. 

Additionally, as for the U.S. legal dataset, \cite{bauer2023legal} proposes a dataset that consists of 430k U.S. court opinions with key passages annotated for the task of legal extractive summarization. The dataset spans from 1755 through 2016 and includes both state and federal court opinions, with each document accompanied by human annotations of key passages that serve as extractive summaries. Another valuable resource is the CaseHOLD dataset, presented in CaseHOLD~\cite{zheng2021does}. It contains over 53,000 multiple-choice questions covering various areas of law, including constitutional law, criminal law, contract law, and tort law. Additionally, the paper introduces two domain pretrained models, BERT-Law and BERT-CaseLaw, which are based on BERT-base but pretrained on different subsets of US legal documents. These models, along with the dataset, contribute to the specialization of LLMs in the legal domain and help address the challenges and limitations of domain pretraining for law.

With regard to the UK legal dataset, \cite{ostling2024cambridge} proposes the Cambridge Law Corpus (CLC) is a comprehensive dataset of over 250,000 UK court cases spanning from the 16th century to the 21st century, created for legal AI research. It includes raw text and metadata, with 638 cases having expert annotations on case outcomes, and is designed to facilitate research in natural language processing and machine learning within the legal domain.

\subsection{Multi-language}

In response to the issue of legal datasets being concentrated in fewer languages (such as English and Chinese),~\cite{niklaus2023multilegalpile} has introduced the MultiLegalPile dataset, a 689GB multilingual legal corpus consisting of 24 languages from 17 jurisdictions, aimed at training large language models (LLMs) in the legal domain.
This expansive dataset serves to enhance the linguistic capabilities and legal understanding of LLMs across diverse legal systems and languages, fostering more comprehensive and inclusive legal AI research and applications.


\section{Conclusion and Future Directions}

In conclusion, integrating large language models (LLMs) into law shows great potential for improving efficiency and accuracy. Despite promising results in tasks like legal document analysis and contract review, concerns regarding privacy, bias, and transparency must be addressed. Moving forward, further research is needed to mitigate biases, enhance interpretability, and develop specialized data resources. Establishing ethical guidelines is crucial for responsible integration. Overall, the use of LLMs holds promise for enhancing legal processes and access to justice.


\bibliographystyle{ACM-Reference-Format}
\bibliography{legalllmsurvey/related_papers}

\end{document}